\documentclass{article}
\usepackage{ijcai18}

\usepackage{times}
\usepackage{xcolor}
\usepackage{soul}
\usepackage[utf8]{inputenc}
\usepackage[small]{caption}
\usepackage{listings}
\usepackage{graphicx}
\usepackage{algorithm}
\usepackage[noend]{algorithmic}
\usepackage{amsmath}
\usepackage{amsfonts}
\usepackage{url}

\newcommand{\argmax}{\mathop{\rm arg~max}\limits}

\title{ClassSim: Similarity between Classes Defined by Misclassification Ratios of Trained Classifiers}

\author{
Kazuma Arino$^1$\thanks{Equal contribution.}, 
Yohei Kikuta$^2$\footnotemark[1]
\\ 
$^1$ Self-employed\thanks{This work was done while at Cookpad Inc.} \\
$^2$ Cookpad Inc. \\
kazuma.arino@gmail.com,
yohei-kikuta@cookpad.com
}

\begin{document}

\maketitle

\begin{abstract}
Deep neural networks (DNNs) have achieved exceptional performances in many tasks, particularly, in supervised classification tasks.
However, achievements with supervised classification tasks are based on large datasets with well-separated classes.
Typically, real-world applications involve wild datasets that include similar classes; 
thus, evaluating similarities between classes and understanding relations among classes are important.
To address this issue, a similarity metric, $ClassSim$, based on the misclassification ratios of trained DNNs is proposed herein. 
We conducted image recognition experiments to demonstrate that the proposed method provides better similarities compared with existing methods and is useful for classification problems\footnote{Source code including all experimental results is available at \url{https://github.com/karino2/ClassSim/}.}.
\end{abstract}

\section{Introduction}
Deep neural networks (DNNs) have demonstrated improved performance for various tasks.
In particular, supervised classification tasks in computer vision are said to be solved.
This statement is correct if the datasets are ideal, i.e., they include a large number of images, well-annotated accurate labels, well-separated, semantically different target classes and identical distributions of training and test data.
As an ideal case, ImageNet~\cite{deng2009imagenet} classes, which are used to evaluate classification tasks, are well-organized~\cite{deselaers2011visual}; usually visually distinct, and distinguishable from a taxonomy perspective.

However, real-world applications typically involve {\it non}-ideal datasets.
For example, consumer generated medias generate huge but wild data~\cite{izadinia2015deep}.
This type of data forms supervised datasets wherein labels are manually assigned by users. 
As a result, in such datasets, labels for given similar images can vary and classes can be disorganized. 
In addition, classes that are objective variables of models are not always well-separated semantically, which means that a dataset may contain similar classes, e.g., spaghetti, carbonara, and alfredo classes.
These classes are similar and difficult to distinguish visually\footnote{Classes also have different granularity. 
However, although this study may be relevant to this issue, it is not considered in this paper.}.

Herein, we focus on the difficulties associated with handling fluctuated labels for given similar images and estimating the similarities between classes. 
Once good similarities are obtained, visual relations among classes are evident and the performance of various machine learning tasks, such as classification, can be improved. 
Note that defining similarities is important, but difficult.
Previous studies have imposed rather strong assumptions, e.g., data probabilistic distributions are Gaussian, simple and low dimensional features can represent various images.

A similarity metric based on the misclassification ratios of a trained DNN is proposed herein. 
The proposed similarity only depends on an assumption that DNN classifiers can capture the characteristics of data distribution. 
We believe this assumption is correct because DNNs, particularly convolutional neural networks, have demonstrated high performance for image classification\footnote{Note that we here ignore fooling images~\cite{nguyen2015deep}, and adversarial examples~\cite{goodfellow2014explaining}, which are created artificially to fool classifiers.}~\cite{russakovsky2015imagenet}. 

We find that the proposed similarity is useful for various vision tasks, such as understanding semantic gaps, creating robust models using misclassified examples~\cite{li2013classifying}, and reorganizing target classes.
To the best of our knowledge, no previous studies have investigated inter-class similarity computations based on DNNs predictions.

\begin{figure*}[t!]
\begin{center}
\includegraphics[width=17.0cm]{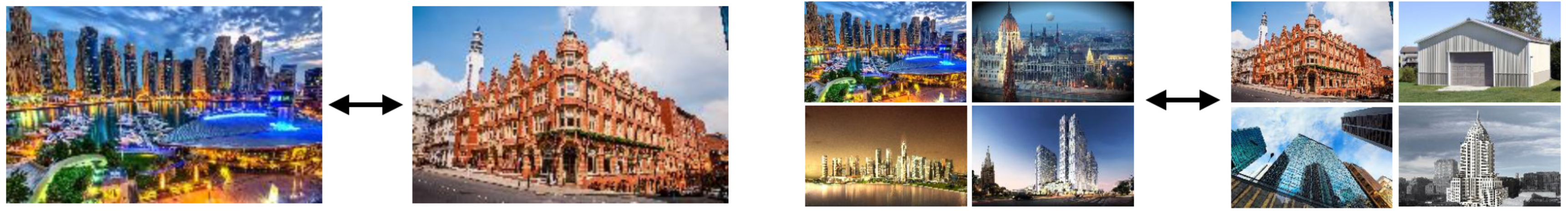}
\caption{Two types of similarities: left, similarity between images; right, similarity between classes (left class is {\it city} and right class is {\it buildings}).
All images in a class have the same labels, and each class may have a different number of images.}
\label{fig:two-types-similarities}
\end{center}
\end{figure*}


\section{Related work}\label{chap:related_work}
There exists two types of similarities in recognition problems; similarity between elements (a pair of images such as single {\it city} image and another single {\it buildings} image) and similarity between classes (a pair of classes such as {\it city} and {\it buildings}), see Figure \ref{fig:two-types-similarities}.
Similarity between elements is employed to search for visually similar products and in visual authentication systems, and similarity between classes is applied to understand semantic gaps and visual taxonomies.

\subsection{Similarities between images}\label{chap:sim-imgs}
Many methods to compute similarity between images have been proposed.
Recently, DNNs have been used to extract image features to compute similarities~\cite{wang2014learning,han2015matchnet}.
For example, DNNs based similarities have been applied to image retrieval~\cite{wu2013online}, person reidentification~\cite{yi2014deep}, facial recognition~\cite{schroff2015facenet}, and visual similarity for product design~\cite{bell2015learning}. 


Note that image-to-image similarity is well studied in various aspects; it is out of the scope of this study.

\subsection{Similarities between classes}\label{chap:sim-cls}
Few methods exist to compute the similarity between classes. 
Compared to image-to-image similarity, estimating class-to-class similarity is much more difficult because a class can include various images and the number of images is not fixed. 

However, a method to estimate the similarities between classes has been proposed~\cite{wang2008measuring,guan2009improved}. 
In that method, images are divided into patches, and features are extracted from each patch using traditional methods, such as RGB color moment.
In addition, to compute the distance between classes, we must assume that the images are generated from Gaussian mixture models (GMMs). 
Note that the number of GMM components must be determined manually relative to the number of target classes. 
In addition, the distance between classes expresses an inverse relation with similarities; they are not normalized, and their absolute values are meaningless.
Here, two distances are involved, i.e., parametric distance (PD), which is the quadratic distance of the means and variances of a GMM, and an approximation of KL divergence.
These two methods return similar results.
Here, strong assumptions and simplifications were used to treat inter-class similarities realistically. 

In this study, we find ways to improve inter-class similarity and compare our results to those obtained using PD.

\subsection{Open set classification}\label{chap:openset-cls}
Open set classification problems~\cite{bendale2015towards} are inherent and difficult in real-world applications.
Thus, few studies have addressed such problems.

However, a solution that employs features extracted using a DNN and meta-recognition has been proposed~\cite{bendale2016towards}.
This solution is useful to eliminate dissimilar unknown unknowns and is, in particular, effective for fooling images.

In addition, support vector machine-based methods have been studied for broader applications. 
Some studies~\cite{scholkopf2001estimating,scheirer2014probability} attempted to discriminate a target class from other classes including unknown unknowns.
Such studies can be interpreted as attempts to find methods to improve one vs. rest (OVR) classifiers to handle unknown unknowns. 
In other words, they attempt to generalize classifiers by isolating a target class from the other classes from various perspectives.
Note that these studies did not employ DNNs.

In this study, as a first step, we created OVR classifiers using a DNN and attempted to improve classification performance using a supervised dataset\footnote{Our original intent was to improve OVR classifiers to handle open set problems in our service.}.

\section{Problem formulation}\label{chap:problem_formulations}

The target problem is defining similarities between classes that include an arbitrary number of images.
Here, let $c \in C$ be a class, $X_c$ be a set comprising images whose labels are all $c$, and $x \in \bigcup_{c} X_c$ be an image. 
The goal is to formulate a quantitative similarity between $c_i$ and $c_j$.
In the following, we consider a case in which one image has one and only one label.

We consider three types of labels. 
The first is latent labels.
We assume that images are generated by unknowable generative models whose latent variables correspond to the labels. 
An image $x$ is generated by following a probabilistic distribution $p(x|c)$. 
Here, any functional form of the distribution is not assumed. 
Generally, latent labels are difficult to estimate by its nature. 
The second is annotated labels.
Here, labels are assigned manually and used as supervised datasets to train a model, corresponding to the labels of $X_c$. 
After being generated, an image from $p(x|c)$ is not always annotated as $c$ owing to stochasticity\footnote{This is natural because annotated labels can differ for different people.
For example, an image generated from {\it buildings} can be annotated as {\it buildings} by one person and {\it city} by another.}. 
We assume that assigning probabilities of annotated labels are controlled by probabilities $p(c|x)$.
The third label type is predicted labels. 
Here, labels are set by the distribution $p(c|x)$ in a deterministic manner. 
A predicted label is determined as follows: 
\begin{eqnarray}
  \argmax_{c \in C} p(c|x) \hspace{0.2cm} \textnormal{where} \hspace{0.2cm} \sum_{c \in C } p(c|x) = 1.
\end{eqnarray}

\begin{figure}[htbp]
\begin{center}
\includegraphics[width=8.5cm]{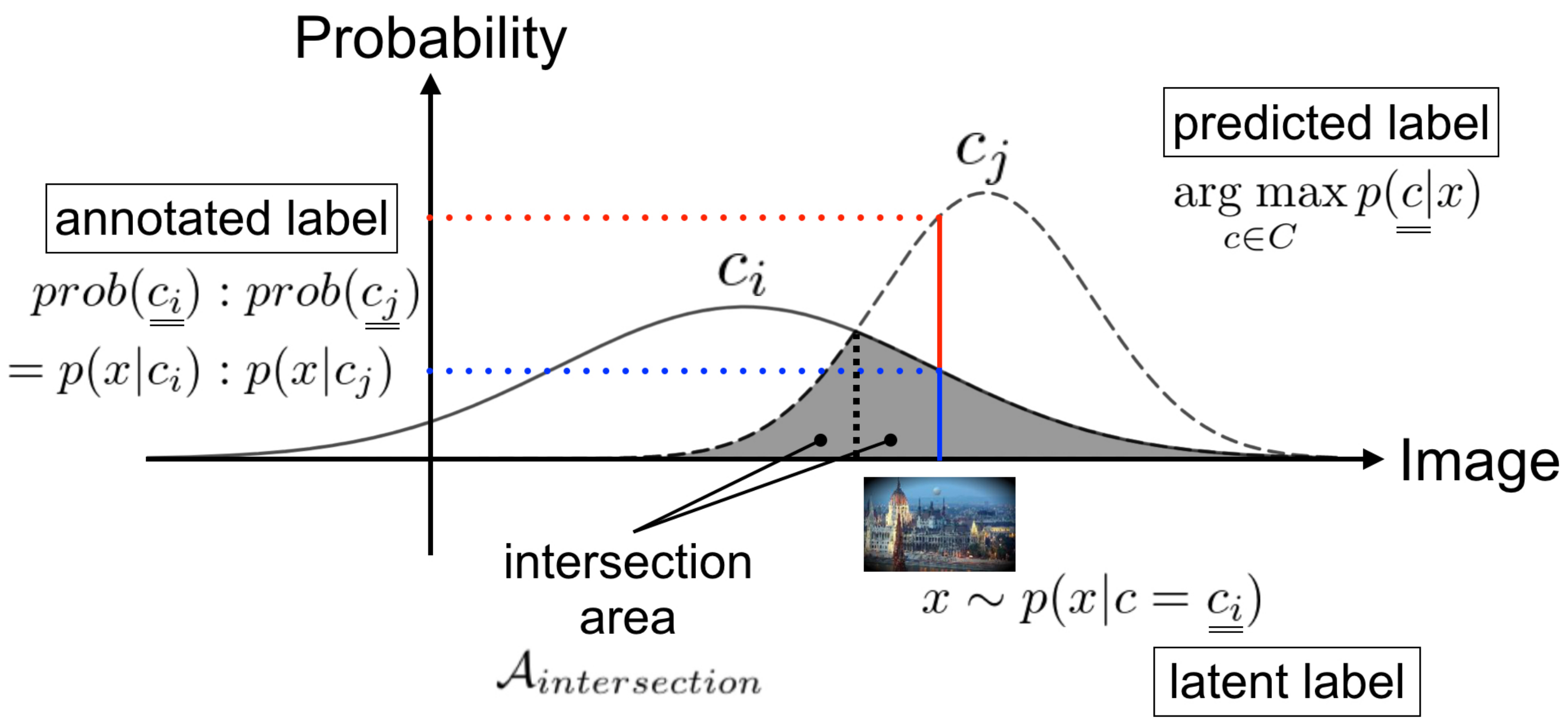}
\caption{Three types of labels. 
The vertical axis is the probability, and the horizontal axis is the image space, which is shown as one dimension for simplicity.
The gray shaded area is the intersection area of the two distributions.}
\label{fig:prob}
\end{center}
\end{figure}

As shown in Figure~\ref{fig:prob}, an image is generated from $p(x|c=c_i)$, labels are assigned by following the probabilities $p(c_i|x)$ and $p(c_j|x)$, and the predicted  label is determined by $\argmax_{c \in C } p(c|x)$.

We define the similarity between classes $c_i$ and $c_j$ as the intersection area of two probability distributions:
\begin{eqnarray}
  {\cal A}_{intersection} = \int_{p(x|c_j) < p(x|c_i)}  p(x|c_j) \; dx  + (i \leftrightarrow j).
  \label{eq:intersection}
\end{eqnarray}
\noindent The intersection area represents the occurrence frequency of a condition wherein it is impossible to uniquely identify the latent labels of the generated images. 
The size of this area reflects the indistinguishability between two classes.
The larger the area, the more similarity the two classes show.
The proposed similarity has the following properties:
\begin{itemize}
\item normalization: possible value range is $[0,1]$,
\item symmetry: $\{c_i, c_j\}$ and $\{c_j, c_i\}$ provide the same value.
\end{itemize}

Since $p(x|c)$ is intractable, an exact computation of similarity is difficult. 
Therefore, the problem is to estimate similarity as approximately as possible using $p(c|x)$, which can be learned approximately from the given datasets $\bigcup_{c} X_c$.

\section{Proposed approach}\label{chap:proposed_approach}
In this section, we propose $ClassSim$ to approximately represent Equation~\ref{eq:intersection}. 
$ClassSim$ is defined using classifiers $f(x)$ trained to learn $p(c|x)$ for the given datasets.
This is the main contribution of this paper.

In addition, we propose $two \ level \ model$ that enhances the performance of OVR classifiers as an application of the proposed $ClassSim$.

\subsection{ClassSim}
We first describe an ideal case. 
Here, the prior distributions are identical for the class pair $(c_i, c_j)$, i.e., $ p(c_i) = p(c_j) $, and we have an ideal binary classifier that returns the score according to the true distribution $p(c|x)$: 

\begin{eqnarray}
  f^{ideal}_{c_j, c_i}(x) = \left\lbrace
\begin{array}{ll}
 1 & \textrm{for} \ \ p(c_j | x) > p(c_i | x), \\
 0 & \textrm{for} \ \ p(c_j | x) \leq p(c_i | x).
\end{array}
\right.
  \label{eq:binary-classfier}
\end{eqnarray}

\noindent For images $ x_{c_i} \in X_{c_i} $ where $c_i$ is the annotated label, misclassification occurs when the classifier returns 1.
Although $f^{ideal}_{c_j, c_i}$ knows the true distribution $p(c|x)$, this misclassification is unavoidable because $x_{c_i}$ can be generated from the region $p(c_j|x) > p(c_i|x)$ (Figure~\ref{fig:prob}).  
 
Let $N_{c_i} = |X_{c_i}|$ and $N_{c_j|c_i}$ be the total number of misclassifications defined as
\begin{eqnarray}
  N_{c_j|c_i} = \sum_{x_{c_i} \in X_{c_i}} I [ f^{ideal}_{c_j, c_i}(x_{c_i}) = 1 ],
  \label{eq:misclassification}
\end{eqnarray}
where $I$ is the indicator function.
Then, we can show the following under ideal conditions:
\begin{eqnarray}
  {\cal A}_{intersection} \simeq \frac{N_{c_j|c_i}}{N_{c_i}}+\frac{N_{c_i|c_j}}{N_{c_j}}.
  \label{eq:approx-area}
\end{eqnarray}

To understand Equation~\ref{eq:approx-area}, consider that the image space is discretized into a finite number of volumes and the distributions remain constant in each volume. 
Then, consider image $x_0 \in X_{c_i}$ satisfying $p(c_j | x_0) > p(c_i | x_0)$ and a small volume $\Delta x$ around the point where the distributions remain constant. 
The effective number within the volume, denoted $\Delta N_{c_j|c_i}$, is expressed as follows:
\begin{eqnarray}
  \Delta N_{c_j|c_i} =  N_{c_i} p(x_0|c_i) \Delta x.
  \label{eq:num-in-volume}
\end{eqnarray}
Taking summation, the left side of Equation~\ref{eq:num-in-volume} becomes
\begin{eqnarray}
  \sum_{\{x | x \in X_{c_i} \bigcap p(c_j | x) > p(c_i | x)\}} \Delta N_{c_j|c_i} = N_{c_j|c_i}.
  \label{eq:sum-lfs}
\end{eqnarray}
The right side of Equation~\ref{eq:num-in-volume} can be expressed as follows: 
\begin{eqnarray}
&& N_{c_i} \sum_{\{x | x \in X_{c_i} \bigcap p(c_j | x) > p(c_i | x)\}} p(x|c_i) \Delta x, \nonumber \\
&=&   N_{c_i} \sum_{\{x | x \in X_{c_i} \bigcap p(x|c_j) > p(x|c_i)\}} p(x|c_i) \Delta x, \\
&\simeq& N_{c_i} \times (\textrm{the right half side of } {\cal A}_{intersection} ),
\end{eqnarray}
where $ \frac{p(c_j | x)}{p(c_i | x)} = \frac{p(x | c_j)}{p(x|c_i)} $, which is ensured by Bayes' theorem and the assumed identical priors. 
By the same argument, by interchanging $i$ and $j$, we can derive Equation~\ref{eq:approx-area}.

\subsubsection{General definition of ClassSim}
Here, we generalize the above ideal binary case. 
Generally, the prior can be different for each class, and the exact form of $p(c|x)$ cannot be obtained. 
Therefore, we define $ClassSim$ constructed by the misclassification ratios of the trained classifiers, which approximate the distribution: 
\begin{eqnarray}
  ClassSim(X_{c_i},X_{c_j}) = \frac{1}{2} \left( \frac{N_{c_j|c_i}}{N_{c_i}} + \frac{N_{c_i|c_j}}{N_{c_j}} \right), 
  \label{eq:classsim}
\end{eqnarray}
where $N_{c_j|c_i}$ is the ratio of the number of elements $x_{c_i} \in X_{c_i}$ predicted as $c_j$ by the classifier. 
Generally, different classifiers can be used to compute $N_{c_j|c_i}$ and $N_{c_i|c_j}$; thus, we require $|C|(|C|-1)$ classifiers to compute the similarities of all pairs of classes in this case. 
The factor 1/2 ensures that the value is in the range [0,1] because the possible maximum value of $N_{c_j|c_i}$ can be $N_{c_i}$. 
This definition obviously possesses symmetry under $i \leftrightarrow j$.

From a classifier perspective, the proposed similarity can be interpreted as the difficulty of classification between two classes.
In addition, scores across different pairs of classes can be compared because their absolute values have meaning, that is, the ratio of misclassification.  

The important points of the proposed similarity are that (1) it only uses trained classifiers and (2) no assumption is made about the functional forms of the distributions or geometric structures of the feature space, which are significant differences observed from previous methods. 
Owing to recent advances in DNN classifiers, it is easier to create good classifiers that can capture the distribution $p(c|x)$ than directly estimating the generative distribution $p(x|c)$.

\subsubsection{One vs. Rest classifier case}
As a concrete classifier case, we introduce an OVR classifiers case for $ClassSim$ computation.
Note that this is one case used for the experiments discussed in the next section. 
In this case, there are $|C|$ classes and $|C|$ classifiers $f_{c,other} \in [0,1]$. 
We can compute $ClassSim$ using $f_{c_i,other}$ and $f_{c_j,other}$.
Here, the number of misclassified samples for $x_{c_i} \in X_{c_i}$ is given by

\begin{eqnarray}
  N_{c_j|c_i} = \sum_{x_{c_i} \in X_{c_i}} I [ f_{c_j, other}(x_{c_i}) > 0.5 ].
  \label{eq:Nij-ovr}
\end{eqnarray}

From an implementation perspective, we only require $|C|$ classifiers to compute the similarities of all pairs rather than $|C| (|C| - 1) /2 $ with the binary classifier case.

Compared to the ideal binary classifier case, we can interpret the OVR classifier $f_{c_j, other}$ as the approximation of $f^{ideal}_{c_j, c_i}$ by averaging $c_i$ for all $ c_i \in C / \{c_j\}$. 
If $c_i$ is similar to $c_j$ and rather different from the other classes, the similarity tends to be large because the classifications are easy to ``misclassify''. 
The misclassification ratio $\frac{N_{c_j|c_i}}{N_{c_i}}$ can be understood as how $c_j$ is similar to $c_i$ compared to the other classes.
From this observation, $ClassSim$ is still a good metric for similarity between two classes.

\subsubsection{Multi-class classifier case}
Here, we consider a multi-class classifier case. 
We require only one classifier $f \in \mathbb{R}_+^{|C|} \ s.t. \ \sum_{c \in C} f_c (x) = 1$ in this case.

The number of misclassified samples for $x_{c_i} \in X_{c_i}$ is given by

\begin{eqnarray}
  N_{c_j|c_i} = \sum_{x_{c_i} \in X_{c_i}} I [ \argmax_c f_{c}(x_{c_i}) = c_j ].
  \label{eq:Nij-multi}
\end{eqnarray}

For a pair of two similar classes, the similarity of the multi-class case shows the same tendency as the OVR classifier case; however, its value is relatively smaller. 
We demonstrate that this phenomenon is true and compare both cases in detail in the next section.

\subsection{Two level model}\label{chap:prop_twolevel}
The proposed $ClassSim$ is useful for understanding the similarities between classes and various applications, such as improving classifiers. 
As an application of $ClassSim$, we introduce $two \ level \ model$ that enhances OVR classifications. 

As stated previously, improvements to OVR classifications lead to better solutions for open set problems. 
Among the many different potential improvement directions, we focus on the classification of datasets that include similar classes because this is a difficult problem in real-world applications for which the proposed similarity has high affinity.

\subsubsection{Baseline model}
The simple OVR classifiers introduced in the previous subsection is used as a baseline model. 
For each target class $c_i$, an OVR classifier $f_{c_i, other}$ is trained using datasets $X_{c_i}$ and $\bigcup_{c \in C/\{c_i\}} X_c$.
In total, we have $|C|$ OVR classifiers. 

In the prediction phase, these trained OVR classifiers are applied in some order.
Here, each OVR classifier is trained individually; thus, the scores across different classifiers cannot be compared. 
Therefore, when the first OVR classifier returning a score above a threshold (we use 0.5 in this paper) is found, we select its target label as a predicted label. 
Although we can use some heuristics based on domain knowledge in practical applications, simple alphabetical order of class names is used herein. 
If no classifier has a score greater than the threshold, the predicted label is defined as {\it none}.

\subsubsection{Two level model}
We propose an enhancement to OVR classifiers by constructing one more set of OVR classifiers $f^{(2)}_{c, other}$ that is applied after the first set of classifiers $f_{c, other}$. 

For each target class $c_i$, $f^{(2)}_{c_i, other}$ is constructed as follows. 
First, a set of classes including similar classes to $c_i$ is defined (we use 0.1 as the similarity threshold in this paper):
\begin{eqnarray}
  C^{sim}_{c_i} = \{ c \in C/\{c_i\} | ClassSim(X_{c_i},X_{c}) > 0.1 \}.
  \label{eq:similar-class}
\end{eqnarray}
Second, OVR classifiers are trained using $X_{c_i}$ and $\bigcup_{c \in C^{sim}_{c_i}} X_c$. 
From the construction procedure considered herein, $f^{(2)}_{c, other}$ can distinguish small differences among similar classes. 
Note that the same threshold can be used for all target classes $c_i$ because $ClassSim$ can compare across different pairs of classes, which is why we can collect similar classes without human intervention.  

Note that a situation in which there is no similar class for some target class may occur.  
In this case, we have no $f^{(2)}_{c, other}$ for the target class. 

$two \ level \ model$ are defined by applying $f^{(2)}_{c, other}$ after performing $f_{c, other}$. 
Here, we require one more threshold for $f^{(2)}_{c, other}$, setting 0.5 as with that of $f_{c, other}$. 
The pseudocode of $two \ level \ model$ is as follows.

\begin{algorithm}
\caption{Definition of $two \ level \ model$}
\label{alg:two-level}
\begin{algorithmic} 
\REQUIRE image $x$, classes $c \in C$, OVR classifiers $f, f^{(2)}$
\FOR{$c \in C$}
\IF{$f_{c, other} (x) > 0.5$}
\IF{ $f^{(2)}_{c, other}$ exists }
\IF{$f^{(2)}_{c, other} (x) > 0.5$}
\RETURN $c$
\ENDIF
\ELSE
\RETURN $c$
\ENDIF
\ENDIF
\ENDFOR
\RETURN {\it none}
\end{algorithmic}
\end{algorithm}

\section{Experiments}\label{chap:exp}
Two experiments were conducted to demonstrate the effectiveness of the proposed methods. 
The first experiment involved estimating the similarities between classes, and the results were compared to those of a previous study~\cite{wang2008measuring}. 
The second experiment focused on enhancing OVR classifiers using $ClassSim$. 

To compare our results with the previous study, we attempted to collect the same datasets (16 classes of images gathered using the Yahoo image search API). 
Unfortunately, this API is no longer available; therefore, we use Bing image search\footnote{https://www.bing.com/?scope=images} to collect nearly equivalent datasets. 
We attempted to collect 1,000 images for each class employed in the previous study, but some of the classes contained less than 1,000 images. 

In total, we obtained (16 classes, 11,803 images). 
We divided these images into (training) : (validation) : (test) = 0.8$\times$0.8 : 0.8$\times$0.2 : 0.2 datasets.

\subsection{Similarities between classes}
We trained 16 OVR classifiers using the training set, and these classifiers were trained using transfer learning from a pre-trained Inception v3 \cite{szegedy2016rethinking}.
We then computed $ClassSim$ on the validation set using the trained classifiers for each pair of classes. 

For comparison, we reproduced the results of the previous study. 
In the previous study, each image was divided into 5$\times$5 patches and traditional image features, such as RGB color moment, were used to compute PDs between classes. 
We used these distance values as similarities (note that smaller values indicate greater similarity). 

In addition, we also conducted the same experiment using a single trained multi-class classifier.
We show computed similarities and differences between the results of the OVR case and those of the multi-class case.

In this subsection we show the three most similar classes for each target class. 
The full results of computed similarities are shown in Appendix \ref{app:full-table}.

\subsubsection{Similar pair}
The results of $ClassSim$ computed by the OVR classifiers and PD are shown in Table \ref{top3}. 

There are some overlaps between the two results. 
For example, the pair ({\it bay}, {\it beach}) was the most similar common pair in both cases, which is a natural result for a human sense. 
In addition, both methods provided \{{\it f-16}, {\it city}, {\it clouds}, {\it bay}\} as the most similar class for \{{\it boeing} and {\it helicopter}, {\it buildings}, {\it sky}, {\it ocean}\}, respectively.

We observed significant differences relative to other combinations. 
For example, the most similar class to {\it city} was {\it buildings} for CS and {\it ocean} for PD. 
This indicates that the proposed method obviously yielded a better result, see Table \ref{classexp}.
Furthermore, CS provided ({\it sunset}, {\it sunrise}) as the most similar pair, whereas PD provided {\it f-16} as the class most similar to {\it sunset} or {\it sunrise}. 
This demonstrates that the proposed method can bridge semantic gaps better than the previous method.

\subsubsection{Comparison within a row}
Here, we compare the relative scores among classes for a single target class, which leads to another advantage of the proposed similarity. 

For example, the three classes most similar to {\it buildings} and its CS scores were \{{\it city}:0.656, {\it ships}:0.092, {\it bay}:0.069\}. 
Here, the score difference of the top two classes (a difference of approximately seven times) seems sensible because {\it city} is similar to {\it buildings} but {\it ships} is not. 
In contrast, PD provided \{{\it city}:9624, {\it bay}:9813, {\it ocean}:10152\}.
Since the score difference between {\it city} and {\it bay} was less than that of {\it bay} and {\it ocean}, PD cannot distinguish as well as the proposed CS. 

As a result, we conclude that the proposed method is much more robust than the previous method.  
In fact, our reproduced results for PD were a little different from those of the original paper.

\subsubsection{Comparison across rows}
Here, we investigate the differences across rows, and focus on {\it birds} and {\it sunrise} for CS. 
The highest score for {\it birds} and {\it sunrise} was 0.045 and 0.902, respectively. 
Note that the latter is more than 20 times greater than the former. 
This result is interpretable because {\it birds} is not similar to any other class and {\it sunrise} is very similar to {\it sunset}. 

However, the same argument cannot be applied for the PD case. 
The shortest distance of {\it birds} was less than that of {\it sunrise}, which indicates that inter-row comparison is clearly meaningless for PD.

In contrast, the proposed method has a clear meaning for its absolute value. 
By definition, the value directly represents the misclassification ratio. 
We can think of the value as a quantitative measure of the challenges in distinguishing two classes. 

Carrying this observation further, we can use the similarity to redesign classes, such as merging similar classes.  
For example, in this case, we may merge {\it bay} and {\it beach} for better classifications.\footnote{We did this kind of redesign target classes in our service and found it effective.}

\subsubsection{Comparison between OVR and multi-class classifiers}
The results of $ClassSim$ computed by the multi-class classifier are shown in Table \ref{top3:ovr-and-multi}, where the results are compared with those of the OVR case.

Overall, the two results show good agreement. 
We can see that both case yielded the same most similar classes for each target class except for \{{\it birds}, {\it city}, {\it ships}\}.
Although there exists other differences in the results, the multi-class case also leads better performances than PD.
We can conclude that the proposed similarity is useful for different types of classifiers. 

Note that the similarities of the multi-class case were lower than those of the OVR case. 
This is a natural consequence because in Equation \ref{eq:Nij-multi}, images whose annotated labels are $c_i$ and predicted labels are $c_j$ are counted as the misclassifications; therefore, images predicted as $c \in C / \{c_i, c_j\}$ do not increase the value of the similarity. 
In contrast, the misclassifications of the OVR case include all images that are predicted as $c_j$ by the binary classifier $f_{c_j, other}$. 

The differences of scores were more obvious for the OVR case than the multi-class case.
For example, the three classes most similar to {\it f-16} and those scores were \{{\it boeing}:0258,\ {\it helicopter}:0.188,\ {\it ships}:0.126\} for the OVR case, and
\{{\it boeing}:0.040, {\it helicopter}:0.038, {\it mountain}:0.013\} for the multi-class case.
The OVR case gave clearer differences between \{{\it boeing},\ {\it helicopter}\}

Let us explain some differences in the results.
The most similar class to {\it city} was {\it buildings} for the OVR case and {\it bay} for the multi-class case. 
This result is reasonable because we found some {\it bay} images contain building. 
The most similar class to {\it ships} was {\it f-16} for the OVR case and {\it bay} for the multi-class case.
In this case it's not easy to judge which result is better. 

We conclude that, in this experiment, $ClassSim$ based on the OVR classifiers is slightly better than that of the multi-class classifier.

\begin{center}
\begin{table*}
\begin{tabular}{|l|l|l|l||l|l|l|}
\hline
& \multicolumn{3}{|c||}{$ClassSim$} & \multicolumn{3}{|c|}{Parametric Distance} \\ \hline
bay &  beach:0.626 & ocean:0.320 & city:0.301 &  beach:6588 & mountain:6951 & birds:7192 \\ \hline
beach & bay:0.626 & ocean:0.245 & mountain:0.114 & bay:6588 & mountain:6909 & birds:7014 \\ \hline
birds & ocean:0.045 & face:0.037 & sunset:0.028 & helicopter:5656 & f-16:6490 & boeing:6490 \\ \hline
boeing & f-16:0.258 & helicopter:0.153 & ocean:0.067 & f-16:3438 & clouds:3525 & helicopter:4918 \\ \hline
buildings &  city:0.656 & ships:0.092 & bay:0.069 & city:9624 & bay:9813 & ocean:10152 \\ \hline
city & buildings:0.656 & bay:0.301 & ships:0.097 & ocean:8576 & bay:8679 & mountain:9585 \\ \hline
clouds  & sky:0.787 & ocean:0.260 & sunset:0.128 & f-16:3421 & boeing:3525 & helicopter:5067 \\ \hline
face & ocean:0.051 & sunrise:0.040 & birds:0.037 & f-16:7768 & helicopter:7849 & clouds:8118 \\ \hline
f-16 & boeing:0.258 & helicopter:0.188 & ships:0.126 & clouds:3421 & boeing:3438 & helicopter:4682 \\ \hline
helicopter & f-16:0.188 & boeing:0.153 & ships:0.098 & f-16:4682 & boeing:4918 & clouds:5067 \\ \hline
mountain & bay:0.188 & beach:0.114 & ocean:0.093 & beach:6909 & bay:6951 & birds:7117 \\ \hline
sky & clouds:0.787 & sunset:0.317 & sunrise:0.302 & clouds:6609 & f-16:7161 & boeing:7467 \\ \hline
ships & f-16:0.126 & ocean:0.108 & helicopter:0.098 & helicopter:7506 & birds:7520 & bay:7983 \\ \hline
sunset & sunrise:0.902 & sky:0.317 & ocean:0.163 & f-16:5253 & boeing:5365 & clouds:5447 \\ \hline
sunrise & sunset:0.902 & sky:0.302 & ocean:0.157 & f-16:5885 & boeing:6028 & clouds:6287 \\ \hline
ocean & bay:0.320 & sky:0.271 & clouds:0.260  & bay:7270 & beach:8070 & mountain:8424 \\ \hline
\end{tabular}
\caption{\label{top3}Top three similar classes and their scores by $ClassSim$ (CS) and Parametric Distance (PD). 
Each row shows the three most similar classes to the class in the first column.
CS is the similarity score ranging from 0 to 1 (higher values indicate greater similarity).
PD is a positive real number (lower values indicate greater similarity).
}
\end{table*}
\end{center}

\begin{center}
\begin{table*}
\begin{tabular}{|l|l|l|l||l|l|l|}
\hline
& \multicolumn{3}{|c||}{$ClassSim$ (OVR)} & \multicolumn{3}{|c|}{$ClassSim$ (multi-class)} \\ \hline
bay & beach:0.626& ocean:0.320& city:0.301& beach:0.246& city:0.123& mountain:0.093 \\ \hline
beach & bay:0.626& ocean:0.245& mountain:0.114& bay:0.246& ocean:0.040& buildings:0.015 \\ \hline
birds & ocean:0.045& face:0.037& sunset:0.028& face:0.011& ocean:0.009& mountain:0.008 \\ \hline
boeing & f-16:0.258& helicopter:0.153& ocean:0.067& f-16:0.040& sky:0.005& helicopter:0.005 \\ \hline
buildings & city:0.656& ships:0.092& bay:0.069& city:0.122& bay:0.044& ships:0.017 \\ \hline
city & buildings:0.656& bay:0.301& ships:0.097& bay:0.123& buildings:0.122& ships:0.013 \\ \hline
clouds & sky:0.787& ocean:0.260& sunset:0.128& sky:0.248& ocean:0.041& mountain:0.021 \\ \hline
face & ocean:0.051& sunrise:0.040& birds:0.037& ocean:0.012& birds:0.011& sunset:0.008 \\ \hline
f-16 & boeing:0.258& helicopter:0.188& ships:0.126& boeing:0.040& helicopter:0.038& mountain:0.013 \\ \hline
helicopter & f-16:0.188& boeing:0.153& ships:0.098& f-16:0.038& ships:0.025& bay:0.011 \\ \hline
mountain & bay:0.188& beach:0.114& ocean:0.093& bay:0.093& clouds:0.021& ocean:0.016 \\ \hline
sky & clouds:0.787& sunset:0.317& sunrise:0.302& clouds:0.248& sunset:0.106& sunrise:0.057 \\ \hline
ships & f-16:0.126& ocean:0.108& helicopter:0.098& bay:0.061& helicopter:0.025& ocean:0.022 \\ \hline
sunset & sunrise:0.902& sky:0.317& ocean:0.163& sunrise:0.353& sky:0.106& ocean:0.026 \\ \hline
sunrise & sunset:0.902& sky:0.302& ocean:0.157& sunset:0.353& sky:0.057& bay:0.020 \\ \hline
ocean & bay:0.320& sky:0.271& clouds:0.260& bay:0.087& clouds:0.041& beach:0.040 \\ \hline
\end{tabular}
\caption{\label{top3:ovr-and-multi}Top three similar classes and their scores by $ClassSim$ computed using the one vs. all (OVR) classifiers and $ClassSim$ computed using the multi-class classifier. 
Each row shows the three most similar classes to the class in the first column. 
The similarity score ranging from 0 to 1 (higher values indicate greater similarity).
}
\end{table*}
\end{center}

\vspace{-2cm}

\begin{center}
\begin{table*}
\setlength{\tabcolsep}{0mm}
\renewcommand{\arraystretch}{0}
\centering
\begin{tabular}{|c|c|c|c|c|c|c|c|c|c|}
\hline
 \includegraphics[width=1.5cm]{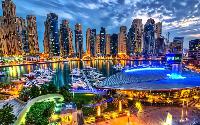} &  \includegraphics[width=1.5cm]{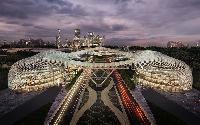} &  \includegraphics[width=1.5cm]{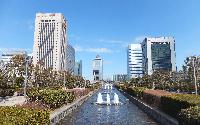} &  \includegraphics[width=1.5cm]{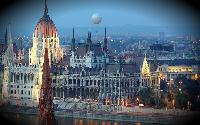} &  \includegraphics[width=1.5cm]{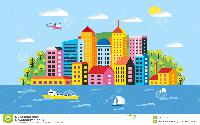} &  \includegraphics[width=1.5cm]{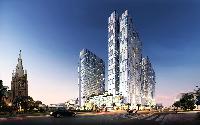} &  \includegraphics[width=1.5cm]{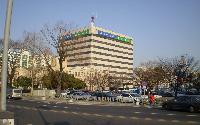} &  \includegraphics[width=1.5cm]{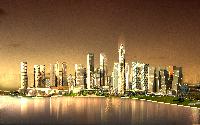} &  \includegraphics[width=1.5cm]{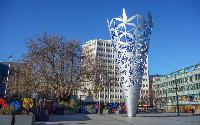} &  \includegraphics[width=1.5cm]{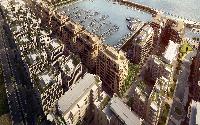}\\ \hline
 \multicolumn{10}{|c|}{city} \\ \hline
\includegraphics[width=1.5cm]{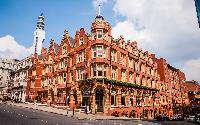} &  \includegraphics[width=1.5cm]{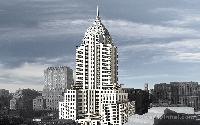} &  \includegraphics[width=1.5cm]{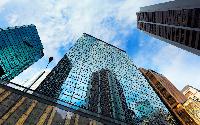} &  \includegraphics[width=1.5cm]{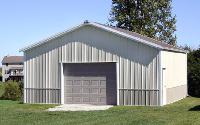} &  \includegraphics[width=1.5cm]{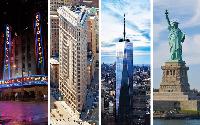} &  \includegraphics[width=1.5cm]{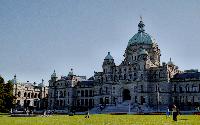} &  \includegraphics[width=1.5cm]{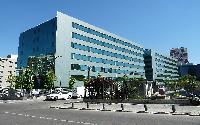} &  \includegraphics[width=1.5cm]{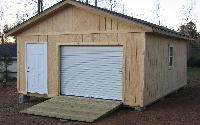} &  \includegraphics[width=1.5cm]{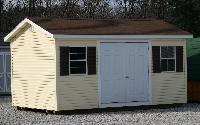} &  \includegraphics[width=1.5cm]{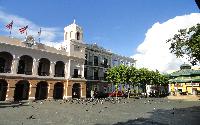}\\ \hline
 \multicolumn{10}{|c|}{buildings} \\[4pt] \hline
\includegraphics[width=1.5cm]{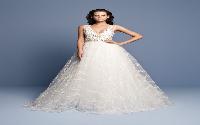} &  \includegraphics[width=1.5cm]{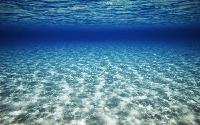} &  \includegraphics[width=1.5cm]{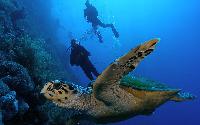} &  \includegraphics[width=1.5cm]{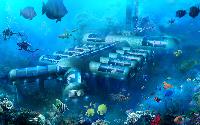} &  \includegraphics[width=1.5cm]{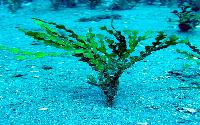} &  \includegraphics[width=1.5cm]{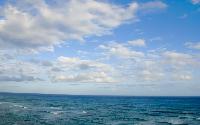} &  \includegraphics[width=1.5cm]{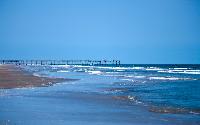} &  \includegraphics[width=1.5cm]{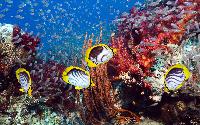} &  \includegraphics[width=1.5cm]{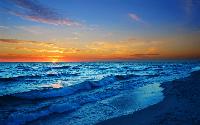} &  \includegraphics[width=1.5cm]{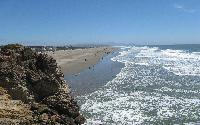}\\ \hline
 \multicolumn{10}{|c|}{ocean} \\[4pt] \hline
\end{tabular}
\caption{\label{classexp} Random samples of images whose classes are {\it city}, {\it buildings}, and {\it ocean}.}
\end{table*}
\end{center}

\vspace{-2cm}

\begin{center}
\begin{table*}
\setlength{\tabcolsep}{0mm}
\renewcommand{\arraystretch}{0}
\centering
\begin{tabular}{|c|c|c|c|c|}
\hline
 \includegraphics[width=3cm]{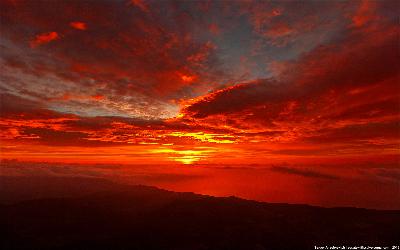} &  \includegraphics[width=3cm]{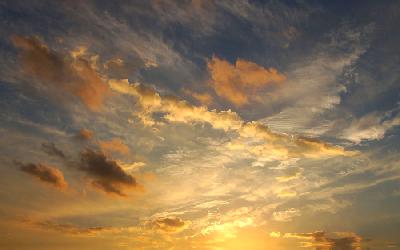} &  \includegraphics[width=3cm]{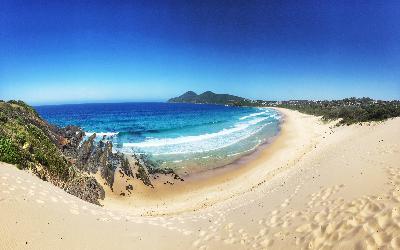} &  \includegraphics[width=3cm]{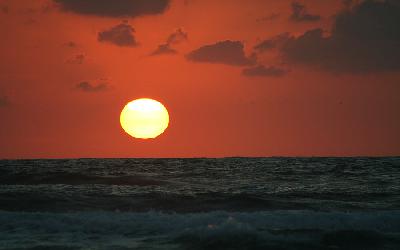} &  \includegraphics[width=3cm]{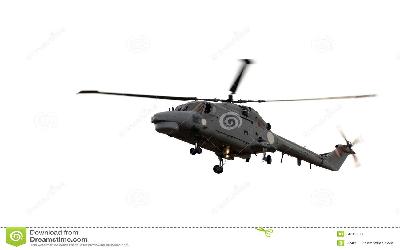} \\ \hline
 sky $\to$ sunrise &  clouds $\to$ sky &  bay $\to$ beach &  ocean $\to$ sunrise &  f-16 $\to$ helicopter \\ \hline
  \includegraphics[width=3cm]{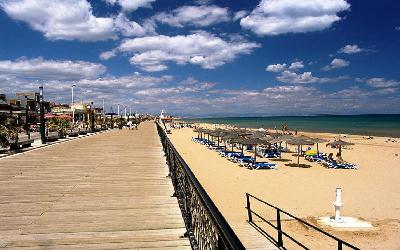} &  \includegraphics[width=3cm]{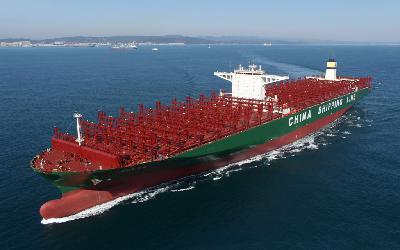} &  \includegraphics[width=3cm]{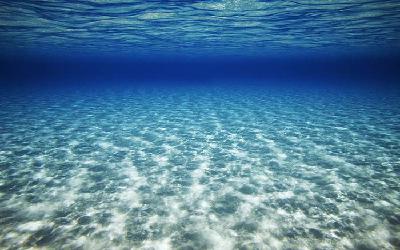} &  \includegraphics[width=3cm]{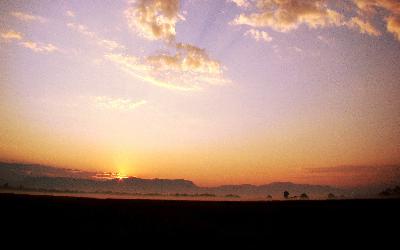} &  \includegraphics[width=3cm]{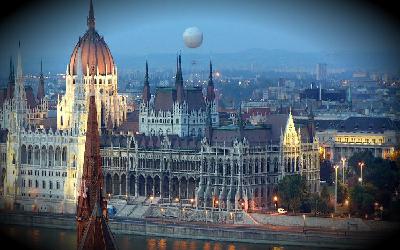} \\ \hline
 bay $\to$ beach &  bay $\to$ ships &  clouds $\to$ ocean &  sky $\to$ sunrise &  bay $\to$ city  \\ \hline
\end{tabular}
\caption{\label{enhance}Images improved by $two \ level \ model$. 
The left class is misclassified by the baseline model. 
The right class is the true label predicted by $two \ level \ model$.
}
\end{table*}
\end{center}

\vspace{2cm}

\subsection{Enhancement of OVR classifiers}
In this experiment, we evaluated the test dataset accuracy of the proposed $two \ level \ model$ by following Algorithm \ref{alg:two-level}. 
Here, we used the same 16 OVR classifiers as in the previous subsection for the first set of classifiers. 
From the results in Table \ref{top3}, we trained 14 $f^{(2)}_{(c, other)}$ with \{training, validation\} datasets because {\it birds} and {\it face} have no similar classes above the threshold. 

The classification results are shown in Table \ref{tb:classification-result}.
The proposed $two \ level \ model$ demonstrated 11\% better accuracy than the baseline model.

\begin{table}[H]
\centering
  \begin{tabular}{|c|c|c|} \hline
     & baseline model & $two \ level \ model$ \\ \hline
    accuracy & 0.552 & \textbf{0.611} \\ \hline
  \end{tabular}
  \caption{Classification results of baseline model and $two \ level \ model$.}
  \label{tb:classification-result}
\end{table}

To observe the ways in which $two \ level \ model$ improved classifications, we show some images in Table \ref{enhance}.
Since $f^{(2)}_{c, other}$ was trained using datasets that only include similar classes, it can distinguish finer differences.

\section{Summary}\label{chap:summary}
Herein,  we formalized the similarities of a pair of classes and proposed $ClassSim$ based on the misclassification ratio of the trained classifiers that can well express the similarities. 

Our experimental results demonstrate that the proposed similarity yields better performance than previous methods. 
The scores were easier to compare across multiple classes, and the differences were much clearer than those of prior studies.
Thus, the proposed method can bridge semantic gaps better than previous methods. 
We then presented the effectiveness of $two \ level \ model$ based on $f^{(2)}_{c, other}$ classifiers trained using only similar classes.
Using the proposed similarity, we could collect similar classes without human intervention.
The experimental results showed that $two \ level \ model$ improved the accuracy of the baseline model that is a simple OVR classi by approximately 11\%.

Note that we have used the model in practical applications with over 150 classes and approximately 500,000 images. 
It has been shown that performance relative to unknown unknowns has been improved. 
In future, we plan to compare the proposed model to previous studies with an open set problem setting comprising publicly available dataset.


\clearpage
\onecolumn
\appendix

\section{Full results of experiments}\label{app:full-table}
In this appendix, we provide the full tables of the similarity computations for both CS(OVR)-PD experiment and CS(OVR)-CS(multi-class) experiment. 
We split the full table into three tables in the both cases.

\subsubsection{CS(OVR) and PD experiment}

\begin{center}
\begin{table*}[h]
\centering
\begin{tabular}{|l|l|l|l|l|l|l|}
\hline
& bay & beach & birds & boeing & buildings & city   \\ \hline
CS & beach:0.626& bay:0.626& ocean:0.045& f-16:0.258& city:0.656& buildings:0.656\\ \cline{0-0}
PD & beach:6588& bay:6588& helicopter:5656& f-16:3438& city:9624& ocean:8576 \\ \hline
CS & ocean:0.320& ocean:0.245& face:0.037& helicopter:0.153& ships:0.092& bay:0.301\\ \cline{0-0}
PD & mountain:6951& mountain:6909& f-16:6490& clouds:3525& bay:9813& bay:8679 \\ \hline
CS & city:0.301& mountain:0.114& sunset:0.028& ocean:0.067& bay:0.069& ships:0.097\\ \cline{0-0}
PD & birds:7192& birds:7014& boeing:6490& helicopter:4918& ocean:10152& mountain:9585 \\ \hline
CS & mountain:0.188& sunrise:0.106& f-16:0.027& ships:0.059& sunset:0.029& beach:0.073\\ \cline{0-0}
PD & ocean:7270& helicopter:7738& sunset:6666& sunset:5365& beach:10429& buildings:9624 \\ \hline
CS & ships:0.077& sunset:0.095& boeing:0.019& city:0.035& beach:0.026& mountain:0.060\\ \cline{0-0}
PD & helicopter:7737& sunset:8008& clouds:6681& sunrise:6028& mountain:10634& beach:9718 \\ \hline
CS & buildings:0.069& city:0.073& sunrise:0.013& bay:0.025& sunrise:0.026& ocean:0.056\\ \cline{0-0}
PD & ships:7983& boeing:8059& beach:7014& birds:6490& ships:10922& birds:10680 \\ \hline
CS & sunset:0.056& sky:0.030& helicopter:0.008& birds:0.019& ocean:0.025& sunrise:0.040\\ \cline{0-0}
PD & city:8679& ocean:8070& mountain:7117& sky:7467& birds:11194& ships:10755 \\ \hline
CS & sunrise:0.055& helicopter:0.030& ships:0.008& face:0.016& mountain:0.020& sunset:0.038\\ \cline{0-0}
PD & boeing:8742& ships:8216& bay:7192& beach:8059& helicopter:12370& helicopter:11876 \\ \hline
CS & sky:0.034& buildings:0.026& city:0.004& sky:0.014& boeing:0.013& boeing:0.035\\ \cline{0-0}
PD & sunset:8842& f-16:8470& ships:7520& face:8123& sunset:13017& sunset:12701 \\ \hline
CS & f-16:0.028& f-16:0.023& sky:0.004& buildings:0.013& sky:0.008& face:0.029\\ \cline{0-0}
PD & f-16:9124& clouds:8854& sunrise:7829& mountain:8323& boeing:13738& boeing:13396 \\ \hline
CS & boeing:0.025& ships:0.023& buildings:0.004& mountain:0.012& helicopter:0.004& f-16:0.018\\ \cline{0-0}
PD & clouds:9547& sunrise:9672& ocean:8899& ships:8564& f-16:13867& f-16:13526 \\ \hline
CS & face:0.007& clouds:0.014& bay:0.004& sunrise:0.011& birds:0.004& helicopter:0.017\\ \cline{0-0}
PD & buildings:9813& city:9718& face:9026& bay:8742& clouds:14416& sunrise:14070 \\ \hline
CS & birds:0.004& boeing:0.011& mountain:0.000& beach:0.011& face:0.004& sky:0.013\\ \cline{0-0}
PD & sunrise:10123& buildings:10429& city:10680& ocean:10683& sunrise:14528& clouds:14118 \\ \hline
CS & helicopter:0.004& face:0.007& clouds:0.000& sunset:0.004& f-16:0.000& birds:0.004\\ \cline{0-0}
PD & face:11249& face:11145& sky:10694& city:13396& face:15108& face:14637 \\ \hline
CS & clouds:0.000& birds:0.000& beach:0.000& clouds:0.000& clouds:0.000& clouds:0.000\\ \cline{0-0}
PD & sky:13553& sky:12997& buildings:11194& buildings:13738& sky:18222& sky:17849 \\ \hline
\end{tabular}
\caption{[1/3] Comparison of the similarities of $ClassSim$ (CS) computed by the one vs. rest (OVR) classifiers and parametric distance (PD). 
Column name represents the target class.
The pairs of \{class : similarity\} are shown in descending order of the similarities for each column.
CS is the similarity score ranging from 0 to 1 (higher values indicate greater similarity).
PD is a positive real number (lower values indicate greater similarity).}
\label{tb:app-ovr-and-pd-1}
\end{table*}
\end{center}

\begin{center}
\begin{table*}[]
\centering
\begin{tabular}{|l|l|l|l|l|l|l|}
\hline
& clouds & face & f-16 & helicopter & mountain & sky \\ \hline
CS & sky:0.787& ocean:0.051& boeing:0.258& f-16:0.188& bay:0.188& clouds:0.787\\ \cline{0-0}
PD & f-16:3421& f-16:7768& clouds:3421& f-16:4682& beach:6909& clouds:6609 \\ \hline
CS & ocean:0.260& sunrise:0.040& helicopter:0.188& boeing:0.153& beach:0.114& sunset:0.317\\ \cline{0-0}
PD & boeing:3525& helicopter:7849& boeing:3438& boeing:4918& bay:6951& f-16:7161 \\ \hline
CS & sunset:0.128& birds:0.037& ships:0.126& ships:0.098& ocean:0.093& sunrise:0.302\\ \cline{0-0}
PD & helicopter:5067& clouds:8118& helicopter:4682& clouds:5067& birds:7117& boeing:7467 \\ \hline
CS & sunrise:0.116& city:0.029& ocean:0.050& beach:0.030& sunrise:0.093& ocean:0.271\\ \cline{0-0}
PD & sunset:5447& boeing:8123& sunset:5253& sunset:5635& helicopter:7604& helicopter:8965 \\ \hline
CS & mountain:0.085& sunset:0.024& sunset:0.030& mountain:0.028& clouds:0.085& mountain:0.055\\ \cline{0-0}
PD & sunrise:6287& sunset:8390& sunrise:5885& birds:5656& sunset:8126& sunset:9274 \\ \hline
CS & beach:0.014& f-16:0.023& bay:0.028& city:0.017& city:0.060& bay:0.034\\ \cline{0-0}
PD & sky:6609& birds:9026& birds:6490& sunrise:6592& ships:8318& sunrise:9310 \\ \hline
CS & ships:0.000& mountain:0.019& birds:0.027& sunrise:0.011& sky:0.055& beach:0.030\\ \cline{0-0}
PD & birds:6681& sunrise:10409& sky:7161& ships:7506& boeing:8323& birds:10694 \\ \hline
CS & helicopter:0.000& boeing:0.016& beach:0.023& birds:0.008& sunset:0.046& boeing:0.014\\ \cline{0-0}
PD & face:8118& mountain:10737& face:7768& mountain:7604& ocean:8424& face:11476 \\ \hline
CS & face:0.000& ships:0.016& face:0.023& face:0.007& helicopter:0.028& city:0.013\\ \cline{0-0}
PD & mountain:8821& beach:11145& beach:8470& bay:7737& f-16:8644& mountain:12548 \\ \hline
CS & f-16:0.000& sky:0.008& city:0.018& sky:0.004& ships:0.020& buildings:0.008\\ \cline{0-0}
PD & beach:8854& bay:11249& mountain:8644& beach:7738& clouds:8821& beach:12997 \\ \hline
CS & city:0.000& helicopter:0.007& sunrise:0.016& ocean:0.004& buildings:0.020& face:0.008\\ \cline{0-0}
PD & ships:9306& sky:11476& ships:8700& face:7849& sunrise:8944& ships:13226 \\ \hline
CS & buildings:0.000& beach:0.007& mountain:0.016& buildings:0.004& face:0.019& f-16:0.005\\ \cline{0-0}
PD & bay:9547& ships:11572& bay:9124& sky:8965& city:9585& bay:13553 \\ \hline
CS & boeing:0.000& bay:0.007& sky:0.005& sunset:0.004& f-16:0.016& helicopter:0.004\\ \cline{0-0}
PD & ocean:11352& ocean:12606& ocean:10834& ocean:9487& buildings:10634& ocean:15269 \\ \hline
CS & birds:0.000& buildings:0.004& clouds:0.000& bay:0.004& boeing:0.012& birds:0.004\\ \cline{0-0}
PD & city:14118& city:14637& city:13526& city:11876& face:10737& city:17849 \\ \hline
CS & bay:0.000& clouds:0.000& buildings:0.000& clouds:0.000& birds:0.000& ships:0.000\\ \cline{0-0}
PD & buildings:14416& buildings:15108& buildings:13867& buildings:12370& sky:12548& buildings:18222 \\ \hline
\end{tabular}
\caption{[2/3] Comparison of the similarities of $ClassSim$ (CS) computed by the one vs. rest (OVR) classifiers and parametric distance (PD). 
Column name represents the target class.
The pairs of \{class : similarity\} are shown in descending order of the similarities for each column.
CS is the similarity score ranging from 0 to 1 (higher values indicate greater similarity).
PD is a positive real number (lower values indicate greater similarity).}
\label{tb:app-ovr-and-pd-2}
\end{table*}
\end{center}

\begin{center}
\begin{table*}[]
\centering
\begin{tabular}{|l|l|l|l|l|}
\hline
& ships & sunset & sunrise & ocean \\ \hline
CS & f-16:0.126& sunrise:0.902& sunset:0.902& bay:0.320\\ \cline{0-0}
PD & helicopter:7506& f-16:5253& f-16:5885& bay:7270 \\ \hline
CS & ocean:0.108& sky:0.317& sky:0.302& sky:0.271\\ \cline{0-0}
PD & birds:7520& boeing:5365& boeing:6028& beach:8070 \\ \hline
CS & helicopter:0.098& ocean:0.163& ocean:0.157& clouds:0.260\\ \cline{0-0}
PD & bay:7983& clouds:5447& clouds:6287& mountain:8424 \\ \hline
CS & city:0.097& clouds:0.128& clouds:0.116& beach:0.245\\ \cline{0-0}
PD & beach:8216& helicopter:5635& sunset:6374& city:8576 \\ \hline
CS & buildings:0.092& beach:0.095& beach:0.106& sunset:0.163\\ \cline{0-0}
PD & mountain:8318& sunrise:6374& helicopter:6592& ships:8780 \\ \hline
CS & bay:0.077& bay:0.056& mountain:0.093& sunrise:0.157\\ \cline{0-0}
PD & boeing:8564& birds:6666& birds:7829& birds:8899 \\ \hline
CS & boeing:0.059& mountain:0.046& bay:0.055& ships:0.108\\ \cline{0-0}
PD & f-16:8700& beach:8008& mountain:8944& helicopter:9487 \\ \hline
CS & beach:0.023& city:0.038& city:0.040& mountain:0.093\\ \cline{0-0}
PD & sunset:8725& mountain:8126& sky:9310& buildings:10152 \\ \hline
CS & sunrise:0.022& f-16:0.030& face:0.040& boeing:0.067\\ \cline{0-0}
PD & ocean:8780& face:8390& beach:9672& sunset:10415 \\ \hline
CS & mountain:0.020& buildings:0.029& buildings:0.026& city:0.056\\ \cline{0-0}
PD & clouds:9306& ships:8725& bay:10123& boeing:10683 \\ \hline
CS & face:0.016& birds:0.028& ships:0.022& face:0.051\\ \cline{0-0}
PD & city:10755& bay:8842& face:10409& f-16:10834 \\ \hline
CS & birds:0.008& face:0.024& f-16:0.016& f-16:0.050\\ \cline{0-0}
PD & buildings:10922& sky:9274& ships:11675& clouds:11352 \\ \hline
CS & sunset:0.004& ships:0.004& birds:0.013& birds:0.045\\ \cline{0-0}
PD & face:11572& ocean:10415& ocean:11797& sunrise:11797 \\ \hline
CS & sky:0.000& helicopter:0.004& helicopter:0.011& buildings:0.025\\ \cline{0-0}
PD & sunrise:11675& city:12701& city:14070& face:12606 \\ \hline
CS & clouds:0.000& boeing:0.004& boeing:0.011& helicopter:0.004\\ \cline{0-0}
PD & sky:13226& buildings:13017& buildings:14528& sky:15269 \\ \hline
\end{tabular}
\caption{[3/3] Comparison of the similarities of $ClassSim$ (CS) computed by the one vs. rest (OVR) classifiers and parametric distance (PD). 
Column name represents the target class.
The pairs of \{class : similarity\} are shown in descending order of the similarities for each column.
CS is the similarity score ranging from 0 to 1 (higher values indicate greater similarity).
PD is a positive real number (lower values indicate greater similarity).}
\label{tb:app-ovr-and-pd-3}
\end{table*}
\end{center}

\clearpage
\subsubsection{CS(OVR) and CS(multi-class) experiment}

\begin{center}
\begin{table*}[h]
\centering
\begin{tabular}{|l|l|l|l|l|l|l|}
\hline
& bay & beach & birds & boeing & buildings & city   \\ \hline
OVR & beach:0.626& bay:0.626& ocean:0.045& f-16:0.258& city:0.656& buildings:0.656\\ \cline{0-0}
multi & beach:0.246& bay:0.246& face:0.011& f-16:0.040& city:0.122& bay:0.123 \\ \hline
OVR & ocean:0.320& ocean:0.245& face:0.037& helicopter:0.153& ships:0.092& bay:0.301\\ \cline{0-0}
multi & city:0.123& ocean:0.040& ocean:0.009& sky:0.005& bay:0.044& buildings:0.122 \\ \hline
OVR & city:0.301& mountain:0.114& sunset:0.028& ocean:0.067& bay:0.069& ships:0.097\\ \cline{0-0}
multi & mountain:0.093& buildings:0.015& mountain:0.008& helicopter:0.005& ships:0.017& ships:0.013 \\ \hline
OVR & mountain:0.188& sunrise:0.106& f-16:0.027& ships:0.059& sunset:0.029& beach:0.073\\ \cline{0-0}
multi & ocean:0.087& sunset:0.014& f-16:0.005& buildings:0.005& beach:0.015& sunset:0.008 \\ \hline
OVR & ships:0.077& sunset:0.095& boeing:0.019& city:0.035& beach:0.026& mountain:0.060\\ \cline{0-0}
multi & ships:0.061& mountain:0.011& boeing:0.005& birds:0.005& ocean:0.013& helicopter:0.008 \\ \hline
OVR & buildings:0.069& city:0.073& sunrise:0.013& bay:0.025& sunrise:0.026& ocean:0.056\\ \cline{0-0}
multi & sky:0.047& sunrise:0.009& clouds:0.005& sunset:0.000& f-16:0.009& mountain:0.008 \\ \hline
OVR & sunset:0.056& sky:0.030& helicopter:0.008& birds:0.019& ocean:0.025& sunrise:0.040\\ \cline{0-0}
multi & buildings:0.044& city:0.008& sunset:0.004& sunrise:0.000& sunrise:0.005& beach:0.008 \\ \hline
OVR & sunrise:0.055& helicopter:0.030& ships:0.008& face:0.016& mountain:0.020& sunset:0.038\\ \cline{0-0}
multi & sunrise:0.020& ships:0.007& sky:0.004& ships:0.000& boeing:0.005& f-16:0.005 \\ \hline
OVR & sky:0.034& buildings:0.026& city:0.004& sky:0.014& boeing:0.013& boeing:0.035\\ \cline{0-0}
multi & sunset:0.013& sky:0.004& ships:0.004& ocean:0.000& sky:0.004& sky:0.004 \\ \hline
OVR & f-16:0.028& f-16:0.023& sky:0.004& buildings:0.013& sky:0.008& face:0.029\\ \cline{0-0}
multi & helicopter:0.011& birds:0.004& beach:0.004& mountain:0.000& mountain:0.004& sunrise:0.000 \\ \hline
OVR & boeing:0.025& ships:0.023& buildings:0.004& mountain:0.012& helicopter:0.004& f-16:0.018\\ \cline{0-0}
multi & f-16:0.009& helicopter:0.000& bay:0.004& face:0.000& sunset:0.000& ocean:0.000 \\ \hline
OVR & face:0.007& clouds:0.014& bay:0.004& sunrise:0.011& birds:0.004& helicopter:0.017\\ \cline{0-0}
multi & face:0.007& face:0.000& sunrise:0.000& clouds:0.000& helicopter:0.000& face:0.000 \\ \hline
OVR & birds:0.004& boeing:0.011& mountain:0.000& beach:0.011& face:0.004& sky:0.013\\ \cline{0-0}
multi & clouds:0.005& f-16:0.000& helicopter:0.000& city:0.000& face:0.000& clouds:0.000 \\ \hline
OVR & helicopter:0.004& face:0.007& clouds:0.000& sunset:0.004& f-16:0.000& birds:0.004\\ \cline{0-0}
multi & birds:0.004& clouds:0.000& city:0.000& beach:0.000& clouds:0.000& boeing:0.000 \\ \hline
OVR & clouds:0.000& birds:0.000& beach:0.000& clouds:0.000& clouds:0.000& clouds:0.000\\ \cline{0-0}
multi & boeing:0.000& boeing:0.000& buildings:0.000& bay:0.000& birds:0.000& birds:0.000 \\ \hline
\end{tabular}
\caption{[1/3] Comparison of the similarities of $ClassSim$ (CS) computed by the one vs. rest (OVR) classifiers and those of CS computed by the multi-class (multi) classifier. 
Column name represents the target class.
The pairs of \{class : similarity\} are shown in descending order of the similarities for each column.
The similarity score ranges from 0 to 1 (higher values indicate greater similarity).
}
\label{tb:app-ovr-and-multi-1}
\end{table*}
\end{center}

\begin{center}
\begin{table*}[]
\centering
\begin{tabular}{|l|l|l|l|l|l|l|}
\hline
& clouds & face & f-16 & helicopter & mountain & sky \\ \hline
OVR & sky:0.787& ocean:0.051& boeing:0.258& f-16:0.188& bay:0.188& clouds:0.787\\ \cline{0-0}
multi & sky:0.248& ocean:0.012& boeing:0.040& f-16:0.038& bay:0.093& clouds:0.248 \\ \hline
OVR & ocean:0.260& sunrise:0.040& helicopter:0.188& boeing:0.153& beach:0.114& sunset:0.317\\ \cline{0-0}
multi & ocean:0.041& birds:0.011& helicopter:0.038& ships:0.025& clouds:0.021& sunset:0.106 \\ \hline
OVR & sunset:0.128& birds:0.037& ships:0.126& ships:0.098& ocean:0.093& sunrise:0.302\\ \cline{0-0}
multi & mountain:0.021& sunset:0.008& mountain:0.013& bay:0.011& ocean:0.016& sunrise:0.057 \\ \hline
OVR & sunrise:0.116& city:0.029& ocean:0.050& beach:0.030& sunrise:0.093& ocean:0.271\\ \cline{0-0}
multi & sunset:0.014& sky:0.008& ships:0.013& city:0.008& f-16:0.013& bay:0.047 \\ \hline
OVR & mountain:0.085& sunset:0.024& sunset:0.030& mountain:0.028& clouds:0.085& mountain:0.055\\ \cline{0-0}
multi & sunrise:0.005& mountain:0.008& buildings:0.009& boeing:0.005& sky:0.013& ocean:0.022 \\ \hline
OVR & beach:0.014& f-16:0.023& bay:0.028& city:0.017& city:0.060& bay:0.034\\ \cline{0-0}
multi & birds:0.005& bay:0.007& ocean:0.009& mountain:0.004& beach:0.011& mountain:0.013 \\ \hline
OVR & ships:0.000& mountain:0.019& birds:0.027& sunrise:0.011& sky:0.055& beach:0.030\\ \cline{0-0}
multi & bay:0.005& sunrise:0.005& bay:0.009& sunset:0.004& city:0.008& face:0.008 \\ \hline
OVR & helicopter:0.000& boeing:0.016& beach:0.023& birds:0.008& sunset:0.046& boeing:0.014\\ \cline{0-0}
multi & ships:0.000& f-16:0.005& face:0.005& sunrise:0.000& birds:0.008& boeing:0.005 \\ \hline
OVR & face:0.000& ships:0.016& face:0.023& face:0.007& helicopter:0.028& city:0.013\\ \cline{0-0}
multi & helicopter:0.000& ships:0.000& city:0.005& sky:0.000& face:0.008& buildings:0.004 \\ \hline
OVR & f-16:0.000& sky:0.008& city:0.018& sky:0.004& ships:0.020& buildings:0.008\\ \cline{0-0}
multi & face:0.000& helicopter:0.000& birds:0.005& ocean:0.000& sunrise:0.005& beach:0.004 \\ \hline
OVR & city:0.000& helicopter:0.007& sunrise:0.016& ocean:0.004& buildings:0.020& face:0.008\\ \cline{0-0}
multi & f-16:0.000& clouds:0.000& sunset:0.000& face:0.000& helicopter:0.004& city:0.004 \\ \hline
OVR & buildings:0.000& beach:0.007& mountain:0.016& buildings:0.004& face:0.019& f-16:0.005\\ \cline{0-0}
multi & city:0.000& city:0.000& sunrise:0.000& clouds:0.000& sunset:0.004& birds:0.004 \\ \hline
OVR & boeing:0.000& bay:0.007& sky:0.005& sunset:0.004& f-16:0.016& helicopter:0.004\\ \cline{0-0}
multi & buildings:0.000& buildings:0.000& sky:0.000& buildings:0.000& buildings:0.004& ships:0.000 \\ \hline
OVR & birds:0.000& buildings:0.004& clouds:0.000& bay:0.004& boeing:0.012& birds:0.004\\ \cline{0-0}
multi & boeing:0.000& boeing:0.000& clouds:0.000& birds:0.000& ships:0.000& helicopter:0.000 \\ \hline
OVR & bay:0.000& clouds:0.000& buildings:0.000& clouds:0.000& birds:0.000& ships:0.000\\ \cline{0-0}
multi & beach:0.000& beach:0.000& beach:0.000& beach:0.000& boeing:0.000& f-16:0.000 \\ \hline
\end{tabular}
\caption{[2/3] Comparison of the similarities of $ClassSim$ (CS) computed by the one vs. rest (OVR) classifiers and those of CS computed by the multi-class (multi) classifier. 
Column name represents the target class.
The pairs of \{class : similarity\} are shown in descending order of the similarities for each column.
The similarity score ranges from 0 to 1 (higher values indicate greater similarity).
}
\label{tb:app-ovr-and-multi-2}
\end{table*}
\end{center}

\begin{center}
\begin{table*}[]
\centering
\begin{tabular}{|l|l|l|l|l|}
\hline
& ships & sunset & sunrise & ocean \\ \hline
OVR & f-16:0.126& sunrise:0.902& sunset:0.902& bay:0.320\\ \cline{0-0}
multi & bay:0.061& sunrise:0.353& sunset:0.353& bay:0.087 \\ \hline
OVR & ocean:0.108& sky:0.317& sky:0.302& sky:0.271\\ \cline{0-0}
multi & helicopter:0.025& sky:0.106& sky:0.057& clouds:0.041 \\ \hline
OVR & helicopter:0.098& ocean:0.163& ocean:0.157& clouds:0.260\\ \cline{0-0}
multi & ocean:0.022& ocean:0.026& bay:0.020& beach:0.040 \\ \hline
OVR & city:0.097& clouds:0.128& clouds:0.116& beach:0.245\\ \cline{0-0}
multi & buildings:0.017& clouds:0.014& ocean:0.014& sunset:0.026 \\ \hline
OVR & buildings:0.092& beach:0.095& beach:0.106& sunset:0.163\\ \cline{0-0}
multi & f-16:0.013& beach:0.014& beach:0.009& sky:0.022 \\ \hline
OVR & bay:0.077& bay:0.056& mountain:0.093& sunrise:0.157\\ \cline{0-0}
multi & city:0.013& bay:0.013& mountain:0.005& ships:0.022 \\ \hline
OVR & boeing:0.059& mountain:0.046& bay:0.055& ships:0.108\\ \cline{0-0}
multi & beach:0.007& face:0.008& face:0.005& mountain:0.016 \\ \hline
OVR & beach:0.023& city:0.038& city:0.040& mountain:0.093\\ \cline{0-0}
multi & birds:0.004& city:0.008& buildings:0.005& sunrise:0.014 \\ \hline
OVR & sunrise:0.022& f-16:0.030& face:0.040& boeing:0.067\\ \cline{0-0}
multi & sunset:0.000& mountain:0.004& clouds:0.005& buildings:0.013 \\ \hline
OVR & mountain:0.020& buildings:0.029& buildings:0.026& city:0.056\\ \cline{0-0}
multi & sunrise:0.000& helicopter:0.004& ships:0.000& face:0.012 \\ \hline
OVR & face:0.016& birds:0.028& ships:0.022& face:0.051\\ \cline{0-0}
multi & sky:0.000& birds:0.004& helicopter:0.000& f-16:0.009 \\ \hline
OVR & birds:0.008& face:0.024& f-16:0.016& f-16:0.050\\ \cline{0-0}
multi & mountain:0.000& ships:0.000& f-16:0.000& birds:0.009 \\ \hline
OVR & sunset:0.004& ships:0.004& birds:0.013& birds:0.045\\ \cline{0-0}
multi & face:0.000& f-16:0.000& city:0.000& helicopter:0.000 \\ \hline
OVR & sky:0.000& helicopter:0.004& helicopter:0.011& buildings:0.025\\ \cline{0-0}
multi & clouds:0.000& buildings:0.000& boeing:0.000& city:0.000 \\ \hline
OVR & clouds:0.000& boeing:0.004& boeing:0.011& helicopter:0.004\\ \cline{0-0}
multi & boeing:0.000& boeing:0.000& birds:0.000& boeing:0.000 \\ \hline
\end{tabular}
\caption{[3/3] Comparison of the similarities of $ClassSim$ (CS) computed by the one vs. rest (OVR) classifiers and those of CS computed by the multi-class (multi) classifier. 
Column name represents the target class.
The pairs of \{class : similarity\} are shown in descending order of the similarities for each column.
The similarity score ranges from 0 to 1 (higher values indicate greater similarity).
}
\label{tb:app-ovr-and-multi-3}
\end{table*}
\end{center}

\clearpage
\twocolumn
\bibliographystyle{named}
\bibliography{ijcai18}

\end{document}